\documentclass[runningheads]{llncs}
\usepackage{graphicx}
\usepackage[a4paper, total={6in, 7in}]{geometry}
\usepackage{csquotes}

\usepackage{wrapfig}
\begin{document}
\title{Automobile Theft Detection by Clustering Owner Driver Data\thanks{This work was supported by Samsung Research Funding \& Incubation Center for Future Technology under Project Number SRFC-TB1403-51.}}
%

\author{Yong Goo Kang\inst{1,2}\and
Kyung Ho Park\inst{2}\and
Huy Kang Kim\inst{2}
}


\authorrunning{Y. G. Kang et al.}
%

\institute{Samsung Research, Samsung Electronics Co. LTD, Seoul, Republic of Korea\\
\email{ygace.kang@samsung.com}
\and
Graduate School of Information Security, Korea University, Seoul, Republic of Korea\\
\email{\{yonggoo,kyungho96,cenda\}@korea.ac.kr}
}

%
\maketitle 
\begin{abstract}
As automobiles become intelligent, automobile theft methods are evolving intelligently. Therefore automobile theft detection has become a major research challenge. Data-mining, biometrics, and additional authentication methods have been proposed to address automobile theft, in previous studies. Among these methods, data-mining can be used to analyze driving characteristics and identify a driver comprehensively. However, it requires a labeled driving dataset to achieve high accuracy. It is impractical to use the actual automobile theft detection system because real theft driving data cannot be collected in advance. Hence, we propose a method to detect an automobile theft attempt using only owner driving data. We cluster the key features of the owner driving data using the k-means algorithm. After reconstructing the driving data into one of these clusters, theft is detected using an error from the original driving data. To validate the proposed models, we tested our actual driving data and obtained 99\% accuracy from the best model. This result demonstrates that our proposed method can detect vehicle theft by using only the car owner's driving data.

\end{abstract}

\section{Introduction}
Automotive technologies have improved the safety and comfort of driving experiences. Existing vehicle systems embed Electronic Control Units (ECUs) to each part of the car, which communicate through a Controller Area Network (CAN). This communication protocol is limited to within a vehicle; thus, existing systems simply share sensor data and control each unit. To overcome this limitation, Vehicle-to-Everything (V2X) technology has been introduced to expand the communication range of a vehicle. By employing more ECUs and modules, modern cars have started to communicate beyond themselves. With In-vehicle Infotainment (IVI), drivers can listen to music or search for the fastest route to their destination. Additional sensors such as cameras, LIDAR, and RADAR have created a much safer driving environment by triggering alerts for dangerous situations. With enormous research effort and investment, cars are becoming more complex, intelligent, and even autonomous.

However, this complexity creates more threats. The increased number of communicating units will generate more vulnerabilities. For instance, previous studies have shown that an entire car system can be infected from indirect physical access to an infotainment system. Not only physical access but also adversaries leverage communication channels such as GPS or satellite radio to intrude the car system \cite{checkoway2011comprehensive}. Various car hacking techniques have been demonstrated in recent years. Charlie Miller and Chris Valasek intruded a car traversing on a highway by leveraging the vulnerabilities of the Chrysler telematics system. They broke into the system and switched on a radio without authentication \cite{reuter}. In the era of advanced V2X communication and autonomous driving, car systems have become more complex. For safe driving, car manufacturers have recognized security concerns related to complicated car systems.

Among the security concerns related to cars, automobile theft is a significant problem. As a car becomes more intelligent, automobile thieves have evolved to become more intelligent as well. Automobile thieves used to require physical access to steal a car. They stole the car by stealing the key or breaking the car window. Currently, automobile thieves operate in a keyless manner, in that they steal a car without any physical access. An increasing number of vehicle thefts through on-board computer hacking has been reported. By exploiting the vulnerabilities of a complex vehicle system, automobile thieves have begun to steal cars remotely \cite{BBC}. Considering actual theft cases in the real world, the industry has started to develop automobile theft detection methods for safe and secure vehicle systems.

To counteract automobile theft, driver identification can be used, in which an owner driver can be authenticated by his/her unique characteristics. Among numerous methods, biometrics have shown promising performance for identifying the owner driver. Biometrics analyze unique features extracted from the human body, such as fingerprints or iris images. Features extracted from the human body are difficult to duplicate; thus, adversaries cannot neutralize an authentication \cite{villa2018survey}. However, hurdles exist when applying biometrics at the industry level. Accuracy is difficult to maintain in biometrics because false authentications may occur. Furthermore, additional modules are required to extract the identification data. Commercializing biometrics will incur additional costs from device installation, thus imposing an economic burden on car manufacturers \cite{nimbhorkar2015survey}. In addition to the information extracted from the human body, academia and industry have started to research alternative data for driver identification.

To avoid the disadvantages of biometrics, driver identification with data-mining methods has been proposed. Data-mining methods analyze the unique driving styles of drivers. As unique as the fingerprint of the human body, each driving style exhibits distinguishable features among different drivers. At a red traffic light, some drivers would step on the brake pedal smoothly multiple times, while others would abruptly step on the brake pedal only once. These driving styles are different among people and are recorded from data generated from the vehicle. By applying numerous algorithms based on supervised learning, previous data-mining methods had efficiently recognized different drivers.

Although data-mining methods with supervised learning showed competent performance, they are difficult to use as an automobile theft-detection method. As supervised learning algorithms require labeled data during training, the thief’s driving data must be provided. However, it is impossible to collect the real thief’s driving data from the crime scene. Currently, in the automobile industry, it is impossible to collect or generate thief data. Without labeled data, supervised learning cannot classify the owner driver from the thief drivers. Furthermore, driver identification methods with supervised learning cannot easily classify newly provided driving styles. As supervised learning performs classification effectively among trained patterns, untrained data will not be classified as a different class. Thus, data-mining methods should be designed to recognize the owner driver correctly and classify any abnormal driving pattern as that of a thief.

To overcome the limitations of previous methods, we propose an automobile theft detection method utilizing unsupervised learning. As unsupervised learning does not necessitate labeled data for training, our model removes the limitations of previous data-mining methods that employed supervised learning. By employing the k-means clustering algorithm, we clustered the owner driver data to create a pool of trusted driving styles. When validation data are provided, our model determines the most similar driving pattern from the pool. If the validation driver was an owner, the driving pattern of the validation data would exist in the owner driver data pool. The driving style is unique and consistent for every driver. Thus, the gap between the validation data and the selected owner driver data from the pool would be small. However, if the validation driver was a thief, the gap between the validation data and the selected owner driver data would become larger. As the thief driver data are not trained in the pool, a significant gap would appear between the validation data and the selected owner driver data. Considering this gap between the validation data and the trained owner driver data, our model classifies the validation driver as the owner or thief. If the gap is larger than a certain threshold, we classify the validation driver as the thief. Based on the presented automobile theft detection method, our contributions are as follows:

1) Practicality: Our model does not require the thief data at the training stage. Thus, it can be applied as an automobile theft detection method in the real world.

2) Accuracy: By ensembling multiple models for theft detection, our model efficiently distinguishes the owner driver from the thief drivers.

3) Economic Feasibility: As our model analyzes the data extracted from the CAN data from the vehicle can be extracted easily without the need for resource-consuming hardware installation.

4) Reality: Our model is trained and validated with the driving data accumulated from actual driving. Therefore, it detects automobile theft effectively in the real world.

\section{Literature Review}

\subsection{Background}

In this section, we provide the preliminaries regarding machine-learning algorithms. Machine learning is a type of artificial intelligence that provides computers with the ability to learn without being explicitly programmed. Machine-learning algorithms are typically classified into three categories, depending on the nature of the data available to learning:

\begin{itemize}
    \item \textit{Supervised learning}: All data are labeled, and the algorithms learn to predict the output from the input data.
    \item \textit{Unsupervised learning}: All data are unlabeled, and the algorithms learn to inherent the structure from the input data.
    \item \textit{Semi-supervised learning}: Some data are labeled but most are unlabeled; a combination of supervised and unsupervised techniques can be used.
\end{itemize}

We used the k-means clustering method in unsupervised learning because we assumed that the thief data were not available for our primary assumption.

\subsection{Related works}
Among numerous driver identification methods, the data-mining method demonstrated higher precision compared with other methods. Based on the training style of the model, previous data-mining methods can be categorized into two groups: supervised learning and unsupervised learning. Past data-mining methods are summarized in Table~\ref{tab:methods}.

\begin{table}
\caption{Data-mining Methods for Driver Identification}
\centering
\begin{tabular}{|c|c|c|c|}
\hline
Training Style & Data Source & Algorithm & Reference\\
\hline
Supervised Learning & Simulator & HMM & \cite{zhang2014study}\\
& Additional Sensor & GMM & \cite{nishiwaki2007driver}\\
& CAN & HMM & \cite{choi2007analysis}\\
& CAN & Machine Learning Algorithms & \cite{enev2016automobile},\cite{kwak2016know},\cite{zhang2019deep}\\
\hline
Unsupervised Learning & Additional Sensor & HCA & \cite{constantinescu2010driving}\\
& Additional Sensor & k-means Clustering & \cite{higgs2014segmentation}\\
\hline
\end{tabular}
\label{tab:methods}
\end{table}

Data-mining with supervised learning utilizes labeled driving data to train the model. First, Zhang \textit{et al.} suggested a driver identification model leveraging the Hidden Markov Model (HMM) to analyze unique driver patterns. From an artificially configured simulator, they extracted features related to an accelerator and steering wheel, and then classified different drivers. The identification model based on the HMM yielded 85\% accuracy, implicating that analyzing the unique characteristics of the driver using the data-mining method was meaningful \cite{zhang2014study}. For an approach more precise to real-world driving, Nishiwaki \textit{et al.} collected actual driving data with a specially designed Toyota Regius that is equipped with additional sensors. They accumulated actual driving data while several drivers drove along city roads. From the collected data, they generated spectral features implicating the different patterns of each driver when the speed is changed. By applying the Gaussian Mixture Model (GMM), their model classified 276 drivers with an accuracy of 76.8\% \cite{nishiwaki2007driver}. Data extracted from additional sensor illustrated the real world better compared with data from a simulator. However, additional hardware installation cost was incurred for data extraction. To apply the proposed model in the industry, economic feasibility must be considered; it is not economical to equip data extraction hardware in all vehicles. Although the model identifies drivers excellently, a lightweight data extraction process is still necessitated.

To satisfy a lightweight data extraction process, some researchers have utilized the driving data extracted from the CAN. As data extraction from the CAN requires only physical access using a single cable, it is more economical than other methods. Choi \textit{et al.} collected CAN data from 9 drivers during actual driving and applied the HMM algorithm to obtain unique driving patterns. The accuracy of driver identification was 25\%, but they showed the possibility of driver identification with CAN data \cite{choi2007analysis}. Enev \textit{et al.} ensembled multiple machine-learning algorithms to enhance the performance of the identification model. They collected actual driving data from 15 different drivers and achieved 100\% identification accuracy \cite{enev2016automobile}. Furthermore, Zhang \textit{et al.} leveraged deep neural networks to scrutinize unique driving patterns. They stacked multiple Convolutional Neural Networks (CNN) and Recurrent Neural Networks (RNN) to illustrate the distribution of driving data, in which excellent driver identification performance was demonstrated \cite{zhang2019deep}. Past studies implied that analyzing CAN data with data-mining methods is effective for recognizing different drivers. However, the proposed models were trained in a supervised manner. For automobile theft detection, we cannot collect and provide labeled thief data to the model. Therefore, data-mining methods not requiring any labeled data should be leveraged. 

As an improvement, some researchers have suggested data-mining methods leveraging unsupervised learning. As the algorithms of unsupervised learning do not require labeled data, they can overcome the limitation of supervised learning. Constantinescu \textit{et al.} used the Hierarchical Cluster Algorithm (HCA) to cluster different driving styles. By embedding additional sensors, they collected features such as speed, altitude, and acceleration from GPS coordinates. Applying Principal Component Analysis (PCA) to the extracted features, they efficiently clustered different driving characteristics \cite{constantinescu2010driving}. Moreover, Higgs and Abbas proposed a driver identification model using the k-means clustering algorithm. By clustering distinct driver patterns, they demonstrated that drivers can be classified in an unsupervised manner \cite{higgs2014segmentation}. In the suggested models, the driver identification method with unsupervised learning yielded meaningful contributions. Nonetheless, a bridge is necessary to create an automobile theft detection method because these models can cluster distinct driving styles, but they cannot recognize which cluster is composed of owner driver data. If the cluster of the owner driver’s driving style cannot be identified, then the model cannot be leveraged as a theft-detection method. Moreover, in the proposed models, driving data extracted from additional sensors are analyzed; thus, they fail to achieve economic feasibility at the industry level.

To satisfy the proposed requirements of automobile theft detection, we propose a detection model leveraging driver identification using a clustering algorithm. By employing the k-means clustering algorithm, we trained only the owner driver data. Subsequently, we designed a model to recognize the owner driver and classify any abnormal driving styles as the thief driver. If abnormal driving style is identified, our model alerts that automobile theft is doubted. In the following sections, we elaborate how the detection model is designed and validate it with actual driving data.

\section{Proposed Methodology}

\begin{figure}[th!] 
\centering
\includegraphics[width=1\linewidth]{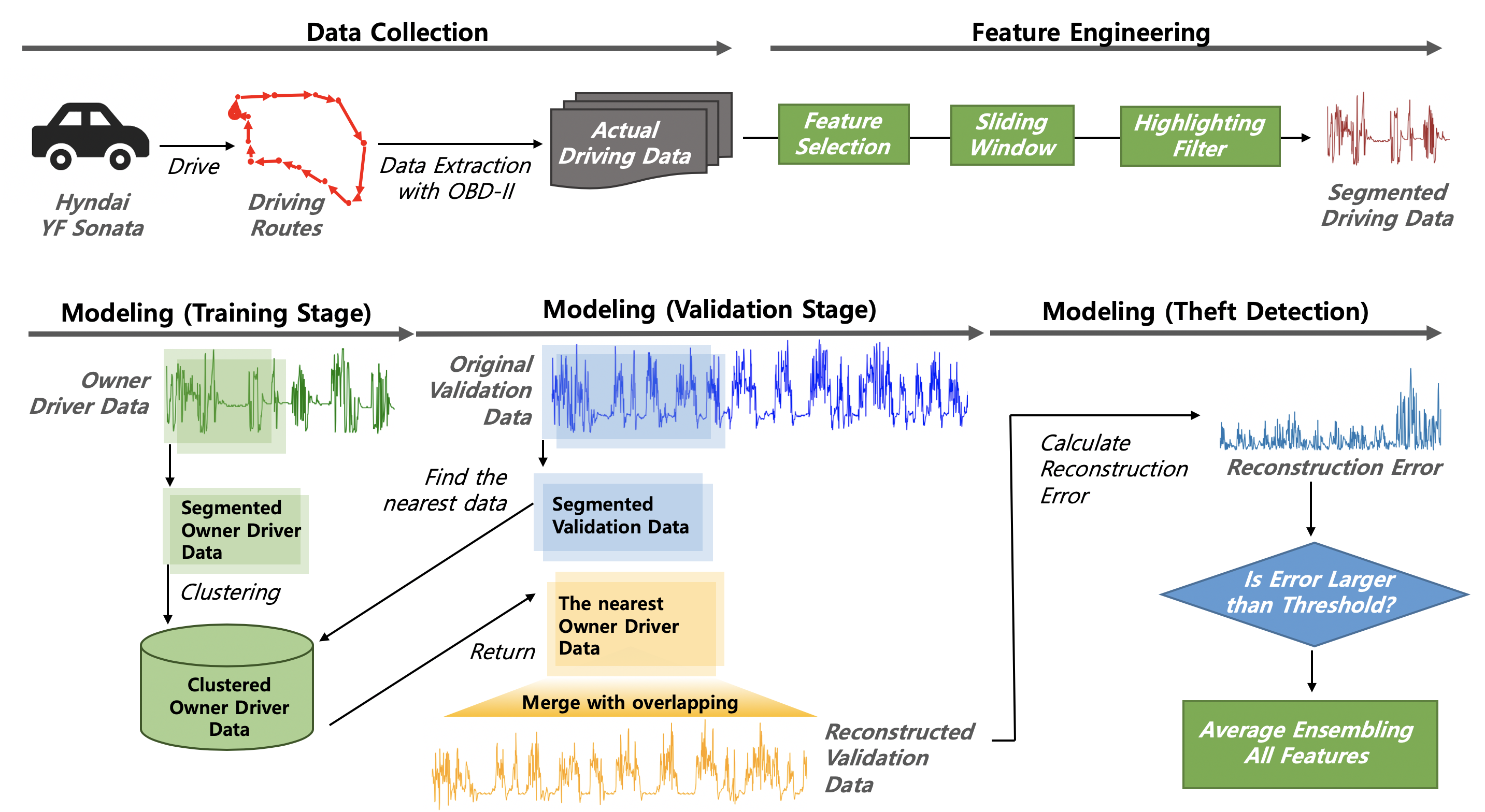}
\caption{Overall Structure of Proposed Methodology} 
\label{fig:overall}
\end{figure}

As illustrated in Fig.~\ref{fig:overall}, our automobile theft detection method comprises three processes, as follows: Data collection, feature engineering, and modeling. In the following, we discuss how we configured the theft detection model using a clustering algorithm.

\subsection{Data Collection}


\begin{wrapfigure}{r}{0.41\textwidth}
\includegraphics[width=1\linewidth]{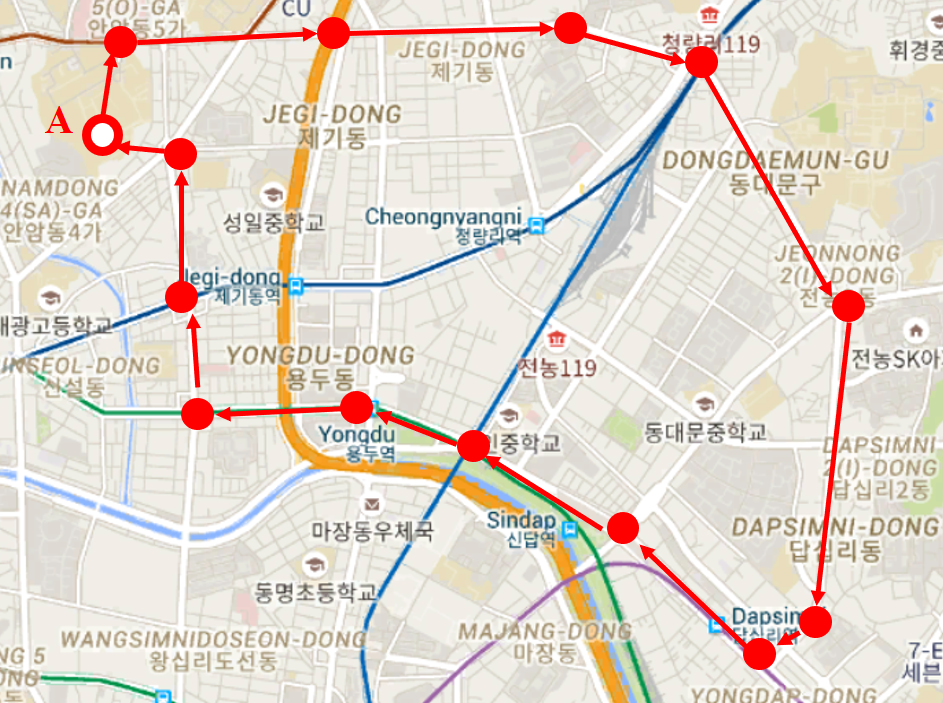}
\caption{Driving Route for Data Collection} 
\label{fig:route}
\end{wrapfigure}

We accumulated actual driving data from four different drivers driving the same route, illustrated in Fig.~\ref{fig:route}. The driving route is circulatory, in the city roads of Seoul, Republic of Korea. The driving route is relatively narrow and includes multiple traffic lights and corners. We formally postulated that the narrow roads and traffic variables would create various driving circumstances; thus, we collected data that fully reflect the driving characteristics of each driver. To minimize noise from data collection, only one vehicle type, i.e., Hyundai YF Sonata, was driven by the drivers at the same time. For data extraction from the CAN, we employed previously proposed data extraction tools \cite{kwak2016know}. We extracted CAN data leveraging On Board Diagnostic-II (OBD-II) and CarbigsP as extraction tools. OBD-II diagnoses each ECU of the vehicle and records its value. Throughout the data collection process, we accumulated 51 features from each driver at each trip. To ensure safety during data collection, all drivers adhered to the traffic rules and regulations of the Republic of Korea.

\subsection{Feature Engineering}
Feature engineering is crucial in our proposed model. It includes four steps for a precise theft detection performance: Feature exploration, feature selection, window sliding, and filter highlighting.

\subsubsection{Feature Exploration}
The raw data extracted from the CAN contain 51 features. To understand the overall configuration of the driving data, we performed feature exploration. As illustrated in Table~\ref{tab:Categories}, we classified the features into three categories based on the data source.

\begin{table}
\caption{Categories of Raw Features}
\centering
\begin{tabular}{|l|l|}
\hline
Category & Description\\
\hline
Fuel & Features related to fuel control, fuel efficiency, and pressure of air inhaled to engine\\
\hline
Engine & Features related to torque, air compressors, and engine coolant temperature\\
\hline
Transmission & Features related to transmission and speed of wheels\\
\hline
\end{tabular}
\label{tab:Categories}
\end{table}

\subsubsection{Feature Selection}
The amount of raw driving data extracted from the CAN was large; thus, feature selection was required to refine the essential features. First, we disregarded non-influential features. If a certain feature satisfies any of the rules described below, we regarded the feature as a non-influential factor for theft detection.

\begin{itemize}
    \item \textit{Rule 1) Missing Value}:Certain feature contains a null value throughout the driving.
    \item \textit{Rule 2) Feature Indifference}: The value of a certain feature is indifferent for each driver.
    \item \textit{Rule 3) Feature Invariance}: Aggregated value of certain feature is zero and standard deviation of that feature is also zero for each driver.
\end{itemize}

\begin{figure}[ht!] 
\centering
\includegraphics[width=14cm]{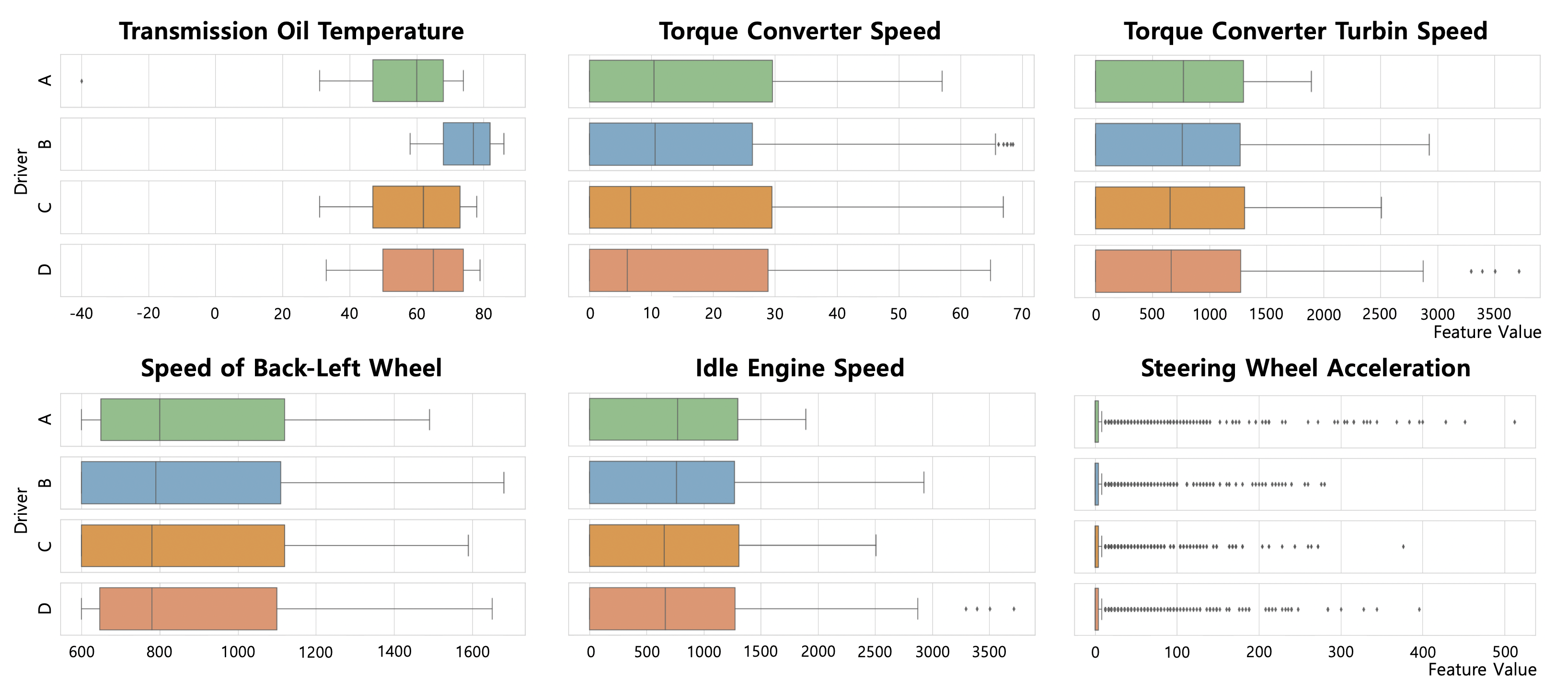}
\caption{Boxplot comparison between essential features and non-essential features}
\label{fig:Boxplots}
\end{figure}

If a feature satisfies the first rule, it implies the existence of data extraction errors. Features containing missing values generate errors during modeling; therefore, we disregarded them. Moreover, Rule 2 proposes that the feature does not imply any distinct characteristics among the drivers. As such a feature degrades the detection model, we eliminated these features based on the second rule. Finally, Rule 3 also implies the existence of data extraction errors. If an unknown error occurs during data extraction, we checked whether OBD-II consistently records a zero value. We disregarded features satisfying Rule 3 to minimize noise from the data extraction error.

We postulated an essential feature as the characteristic that distributes differently among different drivers. After we applied the three feature selection rules, we refined the essential features statistically. The boxplots in Fig.~\ref{fig:Boxplots} vividly show the distinct distribution of the selected features among the drivers. In terms of transmission oil temperature, all statistical characteristics are different among the four drivers. Features such as speed of back-left wheel, torque converter turbine speed, idle engine speed, and torque converter speed shows different maximum values among the drivers. However, the steering wheel acceleration distributes similarly among the four drivers, thus does not imply meaningful characteristics. In our statistical analysis, we selected five essential features as follows: Transmission oil temperature, speed of back-left wheel, torque converter turbine speed, idle engine speed, and torque converter speed.

\subsubsection{Window Sliding}
After selecting the essential features, we performed window sliding to transform the raw data into a trainable form. As the driving time of each driver at each trip was different, we had to fix the length of the driving data. We set 32 s as the window and 16 s as the stride. We fixed the lengths of the window and stride for synchronicity. We generated segmented driving data by the sliding time window with the stride.

\subsubsection{Filter Highlighting}
\begin{figure}[ht!] 
\centering
\includegraphics[width=13cm]{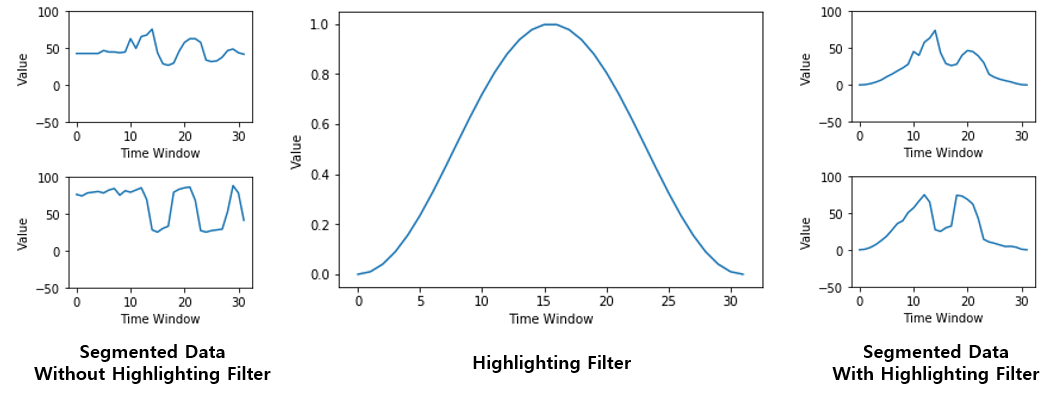}
\caption{Comparison of segmented data with filter highlighting} 
\label{fig:filter}
\end{figure}

To emphasize the unique characteristics of the segmented driving data, we applied filter highlighting. If we simply provide the segmented driving data into the algorithm, the detection performance would be degraded. As shown in Fig.~\ref{fig:filter}, two segmented data without filter highlighting are distributed similarly. As the start and end values of each segments are different, our algorithm classifies two segments as distinct driving styles. Although the two segments show similar driving styles, the clustering algorithm was confusing.

To solve this problem, we applied filter highlighting to the segmented driving data. We utilized a highlighting filter, as illustrated at the center of Fig.~\ref{fig:filter}. It starts at zero and ends at zero, thus multiplying the filter highlighting with the segmented data and transforms the start value and end value to zero. Moreover, it transforms the segmented data to be concentrated in the middle of the time window, emphasizing the unique driving style. The segmented data with filter highlighting are shown at the right side of Fig.~\ref{fig:filter}. After we applied filter highlighting, we found that the algorithm classifies two driving styles in a similar cluster. Therefore, applying filter highlighting effectively reduces algorithm confusion, resulting in more precise theft detection results.

\subsection{Modeling}
The modeling comprises four steps: Clustering, reconstruction, reconstruction error calculation, and theft detection. Clustering is included at the training stage and the others are associated with the validation stage.

\subsubsection{Clustering}
\begin{figure}[ht!] 
\centering
\includegraphics[width=.48\linewidth]{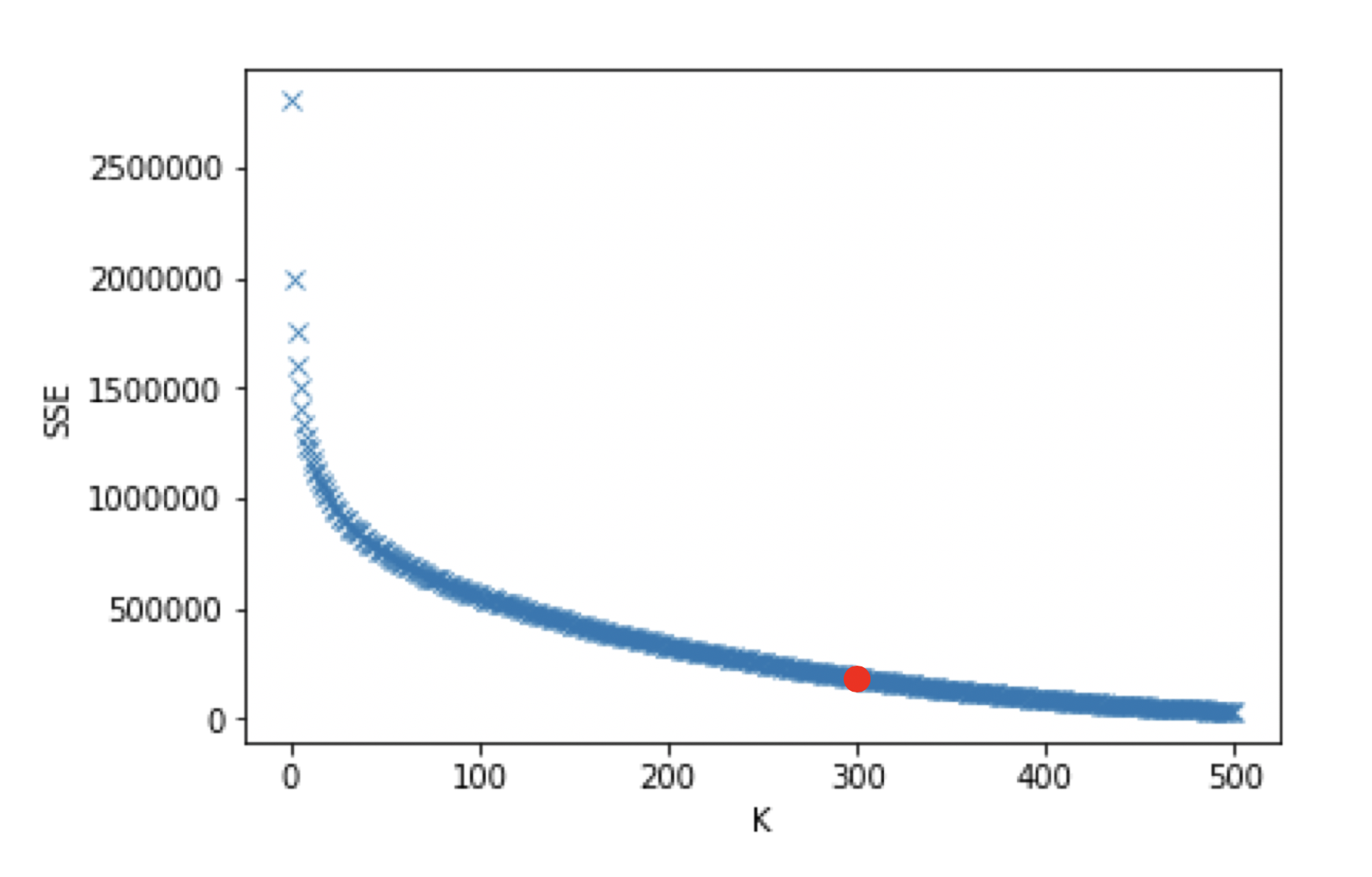}
\caption{Elbow method to obtain optimal size of \textit{k}} 
\label{fig:elbow}
\end{figure}

During the clustering process, we provided the segmented owner driver data to the clustering algorithm. Leveraging the k-means clustering algorithm, we generated clusters of segmented owner driver data as a set of trusted driving patterns. To maximize the theft detection performance, we obtained the optimal size of \textit{k} using the elbow method. The elbow method estimates the Sum of Squared Error (SSE) by increasing the size of \textit{k}. As illustrated in Fig.~\ref{fig:elbow}, the SSE starts to decrease rapidly with the size of \textit{k} but converges at a certain point. We can observe that the applied k-means clustering algorithm is optimized at the size of \textit{k} when the speed becomes slower. After clustering the owner driver data with the optimized size of \textit{k}, we managed the validation data to classify the owner driver from the thieves. In the elbow method, we used 300 as the size of \textit{k} for an accurate performance.

\subsubsection{Reconstruction}
Reconstruction is the process of creating virtual data to show how the validation data are different from the owner driver data. First, we applied feature engineering into the validation data, thus creating segmented validation data. For each segmented validation data, we obtained the nearest owner driver data from the clusters of the owner driver data. Each cluster of the owner driver data can be represented with its central value. When we plotted the segmented validation data into the same space of clusters, a single nearest cluster centroid exists. Setting a distance between the segmented validation data and cluster centroid, the nearest owner driver segment can be obtained. For each segment generated from the validation data, we searched for the nearest owner driver data and merged them by overlapping at a certain length. Consequently, reconstructed validation data of the same form as the original validation data are created.

\subsubsection{Reconstruction Error Calculation}

\begin{figure}[ht!] 
\centering
\includegraphics[width=1\linewidth]{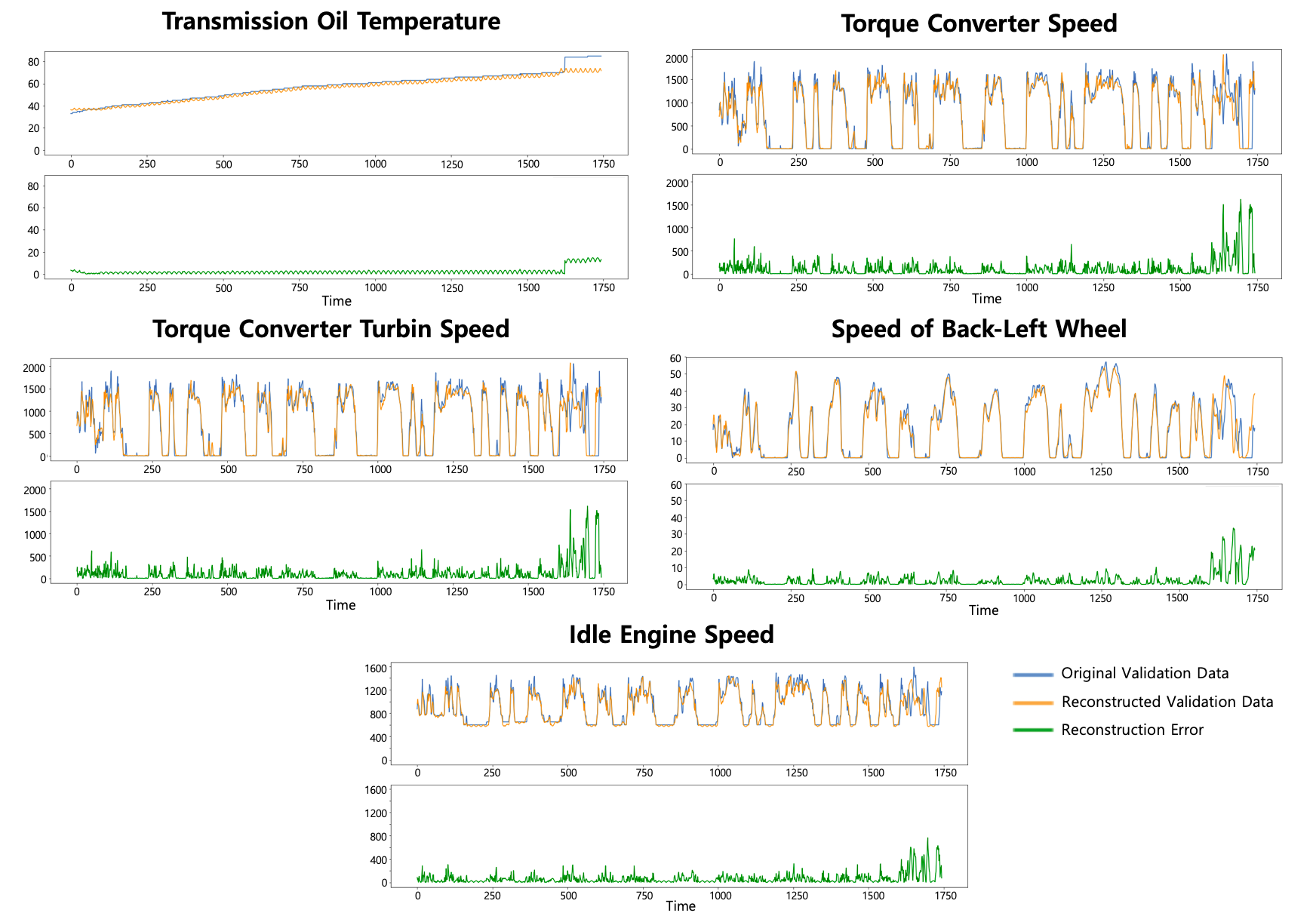}
\caption{Increased reconstruction error with abnormal driving styles} 
\label{fig:proof}
\end{figure}

We set the reconstruction error as a gap between the original validation data and the reconstructed validation data. If the validation driver was the owner driver, the driving style of the validation data would exist in clusters of the owner driver data. When we search for the nearest driving pattern from the clusters, the distance between the validation data and the centroid of the nearest cluster becomes smaller. Thus, the reconstructed validation data would distribute similarly as the original validation data, resulting in a small reconstruction error. However, if the validation driver is a thief driver, the reconstruction error becomes large. As the driving style of the thief driver does not exist in clusters of the owner driver data, the distance between the validation data and the centroid of the nearest cluster becomes larger. This results in reconstructed validation data that are distributed differently from the original validation data. We analyzed the variation in reconstruction error based on whether the validation driver was the owner. Therefore, we set the reconstruction error as a detection criterion for automobile theft detection. To ease theft detection, we calculated the reconstruction error as the absolute value of the difference

To clarify the effectiveness of the reconstruction error as a detection criterion, we proved this concept through simple experiments. As a sample driver, we selected a single trip of randomly selected drivers as the validation driver. To modify the driving data as a stolen case, we substituted a certain length for the data at the end with the other driver’s data. For the selected essential features, we plotted the original validation data, reconstructed the validation data, and reconstruction error at Fig.~\ref{fig:proof}. The reconstructed validation data are distributed in manner similar to the original validation data but are distributed differently at a modified area. This is because abnormal driving styles are not trained in clusters, thus generating large reconstruction errors. We found that reconstruction errors increased excessively at the modified area. By plotting the modified validation data, we discovered different driving styles through the reconstruction process.

\subsubsection{Theft Detection}
We configured five detection models, in which each model analyzes a single feature. We set 32 s as the detection window and calculated the average reconstruction error during the detection window as a representative error. If the representative error is larger than the threshold, we classify the validation data during the detection window as automobile theft. By changing the threshold level, we optimized the individual performances of each model. To maximize the overall detection performance, we performed ensembling using the five detection results predicted by the individual models.

\section{Experimental Results}

With four drivers labeled as driver A, B, C, and D, we selected a single driver as the owner, and the others as thieves. We configured the training set with only the owner driver data and trained the detection model. The validation set was created using the owner driver data from different trips and the thief drivers’ data. We assumed that automobile theft would seldom occur during a vehicle’s lifetime, and thus configured the validation set at the ratio of 8:2 for the owner and thieves, respectively.

\begin{figure}[h!] 
\centering
\includegraphics[width=1\linewidth]{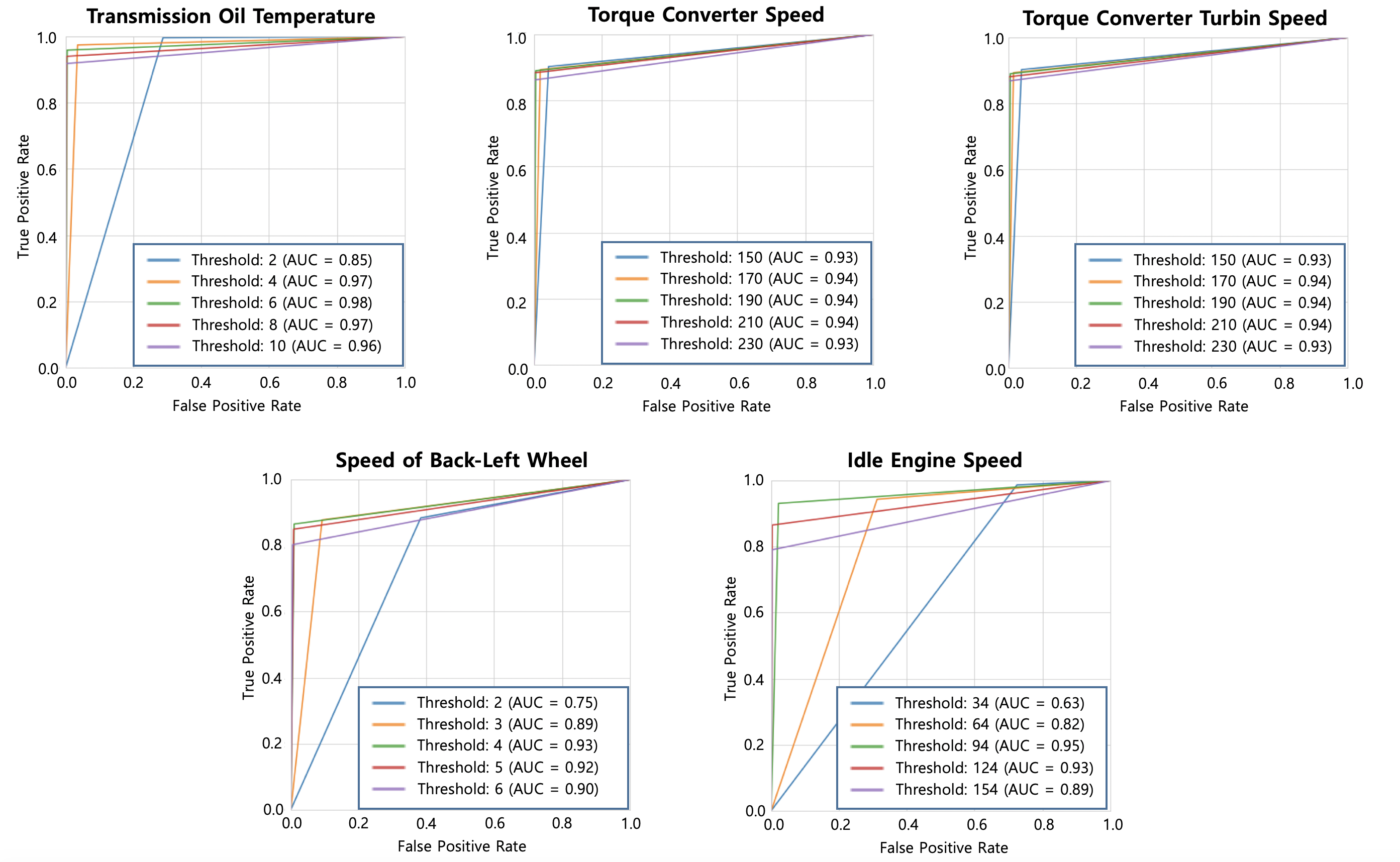}
\caption{ROC curves and AUC values}
\label{fig:roc}
\end{figure}

As illustrated in Fig.~\ref{fig:roc}, we performed iterative experiments while changing the detection threshold value and obtained the optimal threshold values that maximize the AUC value. We individually optimized the detection threshold for each model. We evaluated the performance of the proposed model by measuring several metrics using the optimized threshold values. The metrics used were as follows: accuracy, precision, recall, and F1 score. The experimental results of each individual model are described in Table~\ref{tab:Results}.

\begin{table}
\caption{Experiment Results}
\centering
\begin{tabular}{|c|c|c|c|c|c|c|}
\hline
Model Name & Feature & Optimized Threshold & Accuracy & Precision & Recall & F1 Score\\
\hline
Model TOT & Transmission Oil Temperature & 6 & 0.9901 & 0.9777 & 0.9593 & 0.9684\\
Model TCS & Torque Converter Speed & 190 & 0.9801 & 0.9861 & 0.8875 & 0.9342\\
Model TCTS & Torque Converter Turbin Speed & 190 & 0.9810 & 0.9895 & 0.8875 & 0.9357\\
Model SBW & Speed of Back-Left Wheel & 4 & 0.9706 & 0.9453 & 0.8656 & 0.9037\\
Model IES & Idle Engine Speed & 94 & 0.9826 & 0.9797 & 0.9093 & 0.9432\\
\hline
Model Ensemble & Majority of 5 Models & & 0.9811 & 0.9930 & 0.8875 & 0.9373\\
\hline
\end{tabular}
\label{tab:Results}
\end{table}

Among the features we adopted, \textit{Model TOT} using the transmission oil temperature showed excellent performance in all aspects, i.e., 99.0\% accuracy, 97.8\% precision, 95.9\% recall, and 96.8\% F1 score. Models using other features showed a similar performance, i.e., an accuracy of at least 97\%. For the \textit{Model Ensemble}, the accuracy, recall, and F1 score were close to those of the average of the five models, but the precision was 99.3\% higher than the other models.

\section{Discussion}

\subsubsection{Single and Ensemble Models}
The theft of an automobile can cause serious repercussions once it occurs. Therefore, it is necessary to detect and respond to as many theft situations as possible. In this case, a model with a high recall is necessitated. As shown in Table 3, the accuracy, recall, and F1 Score of \textit{Model TOT} are higher than those of the ensemble model. Therefore, it is preferable to use the single model. In particular, because the recall is 95.9\%, it is possible to detect almost all possible situations of theft. By the individual performances of the five models, we wondered if using only one of the best models would suffice. However, because it is necessary to judge theft by responding to all false alarms, it should be assumed that the human resources and operation cost consumed in detecting and responding to theft would increase. If \textit{Model Ensemble} with high precision is operated simultaneously, it can be used as an auxiliary index to judge theft. The model detects theft as 99.3\% probability of an actual theft occurring. Therefore, when the alarm of \textit{Model Ensemble} is received, the officer and owner can respond more confidently.

\subsubsection{Several Car Owners}
This study assumes that the number of car owners is one. Therefore, the proposed model can be applied only in limited situations such as one household or one driver among family members. In reality, two or three members of a family share a single car is common. In future studies, it is necessary to prove that the number of car owners is more than two and that theft can be detected. In other words, we will construct a model using two or more owner driver data as the training data and verify the model with theft that is not used during learning.

\subsubsection{Learning Algorithms}
Among the clustering algorithms in unsupervised learning, k-means, k-medoids, and DBSCAN algorithms are popularly used. In this study, we used only one of the most popular k-means clustering algorithm. In future studies, we will compare its performance with those of other algorithms. The time complexity of the three algorithms is  \textit{O(n)} for the k-means algorithm, \textit{O($n^{3}$)} for the k-medoids algorithm, and \textit{O(nlogn)} for the DBSCAN algorithm. Therefore, the time performance should also be compared simultaneously.

\section{Conclusion}

Our approach focused on studying a theft detection method applicable to real environments. In this study, we proposed an automobile theft detection method using k-means clustering in unsupervised learning. This method trains the model using only the driving data of the owner driver. Therefore, unlike other methods, our method does not require thief data at the training phase and reflects the actual situation. We used simple CAN-based data that did not require resource-consuming hardware installation. To validate the proposed method, we performed an experiment based on the actual driving data collected from our laboratory. Our experimental results showed that the accuracy of the model for key features was at least 97\% to 99\%, indicating its feasibility as a theft detection solution. Furthermore, this study demonstrated that many situations could be responded to when the theft possibility was based on \textit{Model TOT}, and that cases where theft was confirmed using \textit{Model Ensemble} could be responded to. If our method could detect automobile theft even when two or more car owners were assumed, then it could be commercialized as an automobile theft detection system. In the future, we will study the problems that can occur by applying our method to an automobile theft detection system and the solution to the issues. We hope pertinent industries will apply the proposed method and perform further studies regarding the requirements for using it.

%
%
%
\bibliographystyle{splncs04}
\bibliography{original.bib}

\begin{thebibliography}{10}
\providecommand{\url}[1]{\texttt{#1}}
\providecommand{\urlprefix}{URL }
\providecommand{\doi}[1]{https://doi.org/#1}

\bibitem{checkoway2011comprehensive}
Checkoway, S., McCoy, D., Kantor, B., Anderson, D., Shacham, H., Savage, S.,
  Koscher, K., Czeskis, A., Roesner, F., Kohno, T., et~al.: Comprehensive
  experimental analyses of automotive attack surfaces. In: USENIX Security
  Symposium. vol.~4, pp. 447--462. San Francisco (2011)

\bibitem{choi2007analysis}
Choi, S., Kim, J., Kwak, D., Angkititrakul, P., Hansen, J.H.: Analysis and
  classification of driver behavior using in-vehicle can-bus information. In:
  Biennial workshop on DSP for in-vehicle and mobile systems. pp. 17--19 (2007)

\bibitem{constantinescu2010driving}
Constantinescu, Z., Marinoiu, C., Vladoiu, M.: Driving style analysis using
  data mining techniques. International Journal of Computers Communications \&
  Control  \textbf{5}(5),  654--663 (2010)

\bibitem{BBC}
{Dave Lee}: {Keyless cars 'increasingly targeted by thieves using computers'}.
  \url{https://www.bbc.com/news/technology-29786320} (2014), [Online; Last
  Access 10 June 2019]

\bibitem{enev2016automobile}
Enev, M., Takakuwa, A., Koscher, K., Kohno, T.: Automobile driver
  fingerprinting. Proceedings on Privacy Enhancing Technologies
  \textbf{2016}(1),  34--50 (2016)

\bibitem{higgs2014segmentation}
Higgs, B., Abbas, M.: Segmentation and clustering of car-following behavior:
  Recognition of driving patterns. IEEE Transactions on Intelligent
  Transportation Systems  \textbf{16}(1),  81--90 (2014)

\bibitem{kwak2016know}
Kwak, B.I., Woo, J., Kim, H.K.: Know your master: Driver profiling-based
  anti-theft method. In: 2016 14th Annual Conference on Privacy, Security and
  Trust (PST). pp. 211--218. IEEE (2016)

\bibitem{reuter}
Menn, J.: {Security experts hack into moving car and seize control}.
  \url{https://www.reuters.com/article/us-autos-hacking/security-experts-hack-into-moving-car-and-seize-control-idUSKCN0PV29X20150721}
  (2015), [Online; Last Access 8 May 2019]

\bibitem{nimbhorkar2015survey}
Nimbhorkar, N.M.A.P.S.: A survey paper on continuous authentication by
  multimodal biometric. International Journal of Advanced Research in Computer
  Engineering \& Technology (IJARCET)  \textbf{4}(11) (2015)

\bibitem{nishiwaki2007driver}
Nishiwaki, Y., Ozawa, K., Wakita, T., Miyajima, C., Itou, K., Takeda, K.:
  Driver identification based on spectral analysis of driving behavioral
  signals. In: Advances for In-Vehicle and Mobile Systems, pp. 25--34. Springer
  (2007)

\bibitem{villa2018survey}
Villa, M., Gofman, M., Mitra, S.: Survey of biometric techniques for automotive
  applications. In: Information Technology-New Generations, pp. 475--481.
  Springer (2018)

\bibitem{zhang2019deep}
Zhang, J., Wu, Z., Li, F., Xie, C., Ren, T., Chen, J., Liu, L.: A deep learning
  framework for driving behavior identification on in-vehicle can-bus sensor
  data. Sensors  \textbf{19}(6), ~1356 (2019)

\bibitem{zhang2014study}
Zhang, X., Zhao, X., Rong, J.: A study of individual characteristics of driving
  behavior based on hidden markov model. Sensors \& Transducers
  \textbf{167}(3), ~194 (2014)

\end{thebibliography}

\end{document}